\documentclass{article}

\usepackage[preprint, nonatbib]{neurips_2019}

\usepackage[utf8]{inputenc} 
\usepackage[T1]{fontenc}    
\usepackage{hyperref}       
\usepackage{url}            
\usepackage{booktabs}       
\usepackage{amsfonts}       
\usepackage{nicefrac}       
\usepackage{microtype}      
\usepackage{lipsum}

\usepackage{amsmath}
\usepackage{amsthm}
\usepackage{amssymb}
\usepackage{amsfonts}
\usepackage{mathtools}
\usepackage[dvipsnames]{xcolor}
\usepackage{makecell}
\usepackage{pgfplots}
\usepgfplotslibrary{groupplots}
\usepackage[skip=1pt]{caption}
\usepackage[skip=1pt]{subcaption}
\usepackage{hyperref}
\usetikzlibrary{positioning,shapes}
\definecolor{uzl_red1}{RGB}{181,22,33}
\definecolor{uzl_green}{RGB}{0,75,90}
\definecolor{uzl_green_light}{RGB}{59, 178, 160}
\definecolor{uzl_green2}{RGB}{0, 90, 91}
\definecolor{uzl_blue}{RGB}{0, 70, 114}
\definecolor{uzl_blue_light}{RGB}{60, 169, 213}
\definecolor{uzl_orange}{RGB}{236, 116, 4}

\usepackage[style=numeric,backend=bibtex, maxbibnames=10, firstinits=true,doi=false,isbn=false,url=false]{biblatex}

\tikzset{
	marker style/.style={
		scale=0.35,
		line width=0.7pt,
	},
		marker triangle/.style={
		marker style,
		regular polygon,
		regular polygon sides=3,
	}
}
\newcommand{\markercircle}{\protect\raisebox{0.5pt}{\protect\tikz{\protect\node[draw=uzl_blue, marker style, circle](){};}}}
\newcommand{\markertriangle}{\protect\raisebox{0.5pt}{\protect\tikz{\protect\node[draw=uzl_blue_light, fill=uzl_blue_light, marker triangle](){};}}}
\newcommand{\markerdiamond}{\protect\raisebox{0.5pt}{\protect\tikz{\protect\node[draw=uzl_green_light, fill=uzl_green_light, marker style, diamond](){};}}}

\renewcommand{\vec}[1]{\boldsymbol{\mathrm{#1}}}
\newcommand{\hyp}{\boldsymbol{\theta}}

\title{Scalable Gaussian Process Regression for Kernels with a Non-Stationary Phase}

\author{
  Jan Graßhoff, Alexandra Jankowski, Philipp Rostalski \\
  Institute for Electrical Engineering in Medicine, Universität zu Lübeck\\
  \texttt{j.grasshoff@uni-luebeck.de} \\
}

\begin{document}
\maketitle

\begin{abstract}
The application of Gaussian processes (GPs) to large data sets is limited due to heavy memory and computational requirements. A variety of methods has been proposed to enable scalability, one of which is to exploit structure in the kernel matrix. Previous methods, however, cannot easily deal with non-stationary processes.
This paper presents an efficient GP framework, that extends structured kernel interpolation methods to GPs with a non-stationary phase. We particularly treat mixtures of non-stationary processes, which are commonly used in the context of separation problems e.g.\ in biomedical signal processing. Our approach employs multiple sets of non-equidistant inducing points to account for the non-stationarity and retrieve Toeplitz and Kronecker structure in the kernel matrix allowing for efficient inference. Kernel learning is done by optimizing the marginal likelihood, which can be approximated efficiently using stochastic trace estimation methods. Our approach is demonstrated on numerical examples and large biomedical datasets.
\end{abstract}

\section{Introduction}
Gaussian processes (GPs) provide interpretable models for solving regression, classification and prediction problems in a huge number of scientific domains \cite{Rasmussen_2005}. GP regression originates from early work in geostatistics, where the technique was known under the name of kriging \cite{Rasmussen_2005}. 
Since then, its great potential of discovering intricate structure in data has been shown empirically in numerous problems.  Still, due to $\mathcal{O}(n^3)$ computational and $\mathcal{O}(n^2)$ storage cost in the number of training points~$n$, scalability to large datasets remains as the main limiting factor in many practical applications and restricts GPs to datasets containing at most a few thousand observations \cite{Wilson_2015}.  

Different approaches to scalable GP regression have been proposed \cite{liu2018gaussian}, one of which is based on computing low-rank approximations of the kernel matrix by using sparse inducing point sets \cite{Snelson06,quinonera2005}. Inducing point methods are particularly useful when the data are densely sampled compared to the characteristic length-scale of the underlying process. However, short-scale variability requires a large amount of inducing points, which in such cases diminishes the performance. Also, these methods scale poorly on long time-series data (and spatio-temporal data) as the extending domain has to be filled up with inducing points \cite{Solin2018InfiniteHorizonGP}.

Another orthogonal line of research deals with the exploitation of structure in the kernel matrix: among the most promising approaches are (1)~state-space representation methods \cite{Hartikainen2010KalmanFA, Solin2014} and (2)~methods exploiting Toeplitz and Kronecker matrix structure \cite{Cunningham_2008, Saatci_scalableinference}. The state-space approach enables highly scalable $\mathcal{O}(n)$ inference and marginal likelihood evaluation for stationary time-series data (dimension $D=1$) -- though, it 
might become slow if the gaps between the data points are very uneven. We focus on the second structure exploiting approach, namely Toeplitz/Kronecker matrix structure for fast matrix-vector multiplications (MVMs) which requires that samples are distributed on a multidimensional lattice and that the kernel is stationary/separable. The restriction to lattice structures was later lifted through an approach called structured kernel interpolation (SKI, \cite{Wilson_2015}). It employs a structured set of inducing points and a sparse interpolation matrix enabling fast MVMs with the kernel matrix without requiring any special structure in the data.

In this paper we are concerned with solving the inference and model learning problem for additive mixtures of non-stationary processes.
We show, that SKI can be naturally extended for non-stationary kernels, more specifically kernels with a non-stationary phase in an approach we call warpSKI. We propose to use multiple non-equispaced sets of inducing points to recover structure in the kernel matrix. This permits us to solve new classes of important problems via GPs, in particular temporal/spatio-temporal source separation and regression problems on large biomedical datasets. We also show that the marginal likelihood of non-stationary phase kernels can be efficiently evaluated using recently introduced stochastic trace estimation methods \cite{Dong2017}. The methods in this work are in line with a recent trend in the GP literature to access the kernel matrix only through matrix multiplications, see for instance \cite{Gardner2018, Wang2019}.  We demonstrate scalability of our proposed method to $n\ge 10^5$ points on a numerical example and on openly available biomedical datasets, where the non-stationarity of the process stems from natural fluctuations in the respiratory rate and the heart rate. As in standard SKI, storage complexity is reduced to $\mathcal{O}(n+m)$ and computational complexity to $\mathcal{O}(n+g(m))$ for inference/learning, where $g(m)\leq m\,\log m$ with $m$ being the number of inducing points.
Code implementations of the proposed warpSKI in \textsc{MATLAB} was implemented as an extension to the GPML~4.2 toolbox \cite{Rasmussen_2010} and is available upon request.


\section{Background}
\label{sec:background}

\subsection{Gaussian Process Regression}

This section provides a brief overview of Gaussian processes and the basic computations involved in inference and model learning. The interested reader is referred e.g.\ to \cite{Rasmussen_2005} for more details.  A Gaussian process $f(\vec{x})\sim\mathcal{GP}(0,k_f(\vec{x},\vec{x}'))$ with $\vec{x} \in \mathbb{R}^D$ encodes the prior belief in the distribution of the function values $f(\vec{x})$ and is fully specified by a kernel function $k_f(\vec{x},\vec{x}')$, which is parameterized in a (usually low) number of hyperparameters $\hyp$. The choice of the kernel allows to define the properties of the function $f$, e.g.\ its noise color or periodicity. A GP can formally be understood as an infinite-dimensional generalization of the normal distribution. 
For any finite set of points $X=\{ \vec{x}_1, \dots, \vec{x}_n \}\subset\mathbb{R}^D$, the function $f(\vec{x}_i)$ evaluated at those points has a multivariate Gaussian distribution
\begin{equation*}
    \vec{f} =\left[f(\vec{x}_1), \dots f(\vec{x}_n) \right]^\top \sim \mathcal{N}(\vec{0}, K_{f,XX}),
\end{equation*}
where the entries of the kernel matrix ${(K_{f,XX})}_{i,j}$ are formed by evaluating the covariance function for all pairs of inputs $\vec{x}_i$ and $\vec{x}_j$. In most GP applications it is assumed, that only noisy measurements of the latent function values are available, which we denote as a vector of targets $\vec{y} = ( y_1, \dots, y_n )^\top \in \mathbb{R}^{n}$. 
Here we assume that the measurements are subject to additive noise $y = f(\vec{x}) + \epsilon(\vec{x})$ which is in turn a Gaussian process $\epsilon\sim\mathcal{GP}(0,k_\epsilon(\vec{x},\vec{x}'))$.
In this specific case, the predictive distribution at $n_*$ test points $X_*$ has a closed form solution given by
\begin{equation}
    \begin{split}
   \vec{f}_*\vert & X,X_*,\vec{y},\hyp,\sigma^2 \sim \mathcal{N}(\overline{\vec{f}}_*, \mathrm{cov}(\vec{f}_*)), \\
    \overline{\vec{f}}_* &= K_{f,X_*X}[K_{f,XX} + K_{\epsilon, XX}]^{-1}\vec{y}, \\
    \mathrm{cov}(\vec{f}_*) &= K_{f,X_*X_*} - K_{f,X_*X}[K_{f,XX} +  K_{\epsilon, X X}]^{-1}K_{f,XX_*},
\end{split}
\label{eq:standard_GPR}
\end{equation}
where $\vec{f}_*$ is the vector of function values evaluated at the test points and matrices of the form $K_{X_iX_j}$ denote cross-covariances between respectively two sets of points $X_i$ and $X_j$. In the standard GP regression setting, $\epsilon$ is assumed to be a Gaussian white noise process with variance $\sigma^2$, which corresponds to choosing $K_{\epsilon, X X} = \sigma^2 I$.

The hyperparameters $\hyp$ of the kernel are usually learned directly from the data by optimizing the negative log marginal likelihood 
\begin{equation}
-\mathrm{log}\,\mathcal{L}(\hyp\vert\vec{y}) \propto  \boldsymbol{\mathrm{y}}^\top (K_{f,XX} +  K_{\epsilon, XX})^{-1} \boldsymbol{\mathrm{y}} + \mathrm{log} \, |K_{f,XX} +  K_{\epsilon,XX}|,
\label{eq:log_marg_lik}
\end{equation}
which can be done e.g.\ by gradient-based minimization or sampling. 
Computing the inverse and the log determinant of $K_{f,XX} +  K_{\epsilon,XX}$ are the main bottlenecks for GP inference and model learning. Both involve computing the Cholesky factorization which leads to an overall complexity of $\mathcal{O}(n^3)$. The storage complexity is determined by the need to store the full kernel matrix, leading to $\mathcal{O}(n^2)$.

\subsection{Kronecker and Toeplitz Methods}
There is a growing line of research investigating the exploitation of structure in the kernel matrix to achieve scalable GP inference and model learning.  
When input points are on a multidimensional rectilinear lattice (not necessarily equispaced) and the kernel is separable along input dimensions,  $k(\vec{x},\vec{x}') = \prod_{d=1}^D k^{(d)}(\vec{x}^{(d)},\vec{x}'^{(d)})$, the kernel matrix has Kronecker structure, i.e.\ it can be written as $K = K_1 \otimes \dots \otimes K_D$. This enables fast MVMs (in $\mathcal{O}(D n ^{\frac{D+1}{D}})$ time \cite{Wilson_2014}) and the eigendecomposition of the full matrix can be efficiently calculated by separately taking the eigendecompositions of the smaller matrices $K_i$. Thus, in the GP regression setting subject to Gaussian white noise, the solution to the linear system $(K_{X X} + \sigma^2 I)^{-1} \vec{y}$ and the log determinant $\mathrm{log} \, |K_{X X} + \sigma^{2}I|$ can be evaluated efficiently. Given the eigendecomosition $K=QVQ^\top$, one can use $(K_{X X} + \sigma^2 I)^{-1} \vec{y} = Q(V+\sigma^2 I)^{-1}Q^\top\vec{y}$, where the inversion is trivial and $Q$ can be written as a Kronecker product \cite{Wilson_2015}. Also, the log determinant can be determined based on the eigendecomposition, using $\mathrm{log} \, |K_{X X} + \sigma^{2}I| = \sum_i \mathrm{log}(V_{ii} + \sigma^2)$ \cite{Wilson_2015}.

In \cite{Cunningham_2008} another, orthogonal method is proposed to exploit Toeplitz structure (constant diagonals of the kernel matrix), which arises when the input points are placed equidistantly in $\mathbb{R}$ and the kernel is stationary (that is, $k_i(x,x') = k_i(\tau)$, with $\tau = x-x'$). Toeplitz structure allows for fast MVMs using Fourier transforms \cite{Cunningham_2008, Wilson_2015}, thus the matrix inverse can be computed by conjugate gradients in $\mathcal{O}(n \, \mathrm{log}\,n)$. 
Kronecker and Toeplitz methods complement each other in the sense that they exploit multidimensional and 1D structure, respectively. The Appendix contains a comprehensive overview of fast Kronecker/Toeplitz methods in the GP setting.




\subsection{Structured Kernel Interpolation}
Kronecker and Toeplitz methods are restricted to a few highly specialized problems due to the requirement, that input points are either equispaced or on a multidimensional lattice. The highest performance gains are reached when both of these requirements are met.
Wilson et al.~\cite{Wilson_2015} presented a general purpose inference framework, that exploits structure even for partial-grid/unstructured data by placing inducing points on a multidimensional equispaced lattice. Inducing point methods have long been used throughout the literature for large-scale GP applications and reduce runtime cost to $\mathcal{O}(m^3+m^2n)$ and storage cost to $\mathcal{O}(mn + m^2)$ \cite{quinonera2005}, where $m$ is the number of inducing points. Usually inducing point methods perform best when the data are densely sampled and few inducing points suffice, i.e.\ $m\ll n$. One of the most prominent methods is the subset of regressors (SoR) \cite{Silverman85}, which uses the following low-rank kernel approximation:
\begin{equation}
    k_{\mathrm{SoR}}(\vec{x},\vec{x}') = K_{\vec{x} U} K_{U U}^{-1} K_{U \vec{x}'},
    \label{eq:sor}
\end{equation}
where $U$ is a set of inducing points and $K_{\vec{x} U}$, $K_{U U}$ and $K_{U \vec{x}'}$ are the exact kernel matrices with dimensions $1\times m$, $m\times m$ and $m\times 1$.
Still, inducing point methods suffer from the possibly high amount of inducing points $m$ and the need to sample the full input domain, which is in particular detrimental in temporal/spatio-temporal regression tasks.
Wilson et al.~\cite{Wilson_2015} introduced an approximation called structured kernel interpolation (SKI) of the form $\tilde{K}_{X U} = W K_{U U}$ using an interpolation weight matrix $W\in\mathbb{R}^{n\times m}$. The interpolation matrix can be made very sparse, consisting of only four non-zero entries per row determined by local cubic interpolation. The full approximate kernel matrix can thus be written as
\begin{equation}
    K_{X X} \approx W K_{U U} W^\top \coloneqq K_{\mathrm{SKI}},
\end{equation}
which becomes clear by insertion of $\tilde{K}_{X U}$ in (\ref{eq:sor}). MVMs with $W$ can be computed in $\mathcal{O}(n)$ time and, when placing the inducing points on a lattice, MVMs with $K_{U U}$ can exploit Kronecker and Toeplitz structure, with worst-case cost of $\mathcal{O}(m\log{m})$ for only Toeplitz structure -- the total runtime for MVMs is thus $\mathcal{O}(n + m\log{m})$ \cite{Wilson_2015}. When both Toeplitz structure and Kronecker structure are exploited, the total runtime for MVMs becomes $\mathcal{O}(n + g(m))$, where $g(m)<m\log{m}$ and in many cases we can assume a quasi-linear complexity $g(m)\approx m$, \cite{Wang2019}. Storage costs are reduced to $\mathcal{O}(n+m)$. Thus, SKI significantly relaxes restrictions on the number of inducing points, allowing even for $m\approx n$.


\section{Scalable GPs with a Non-Stationary Phase}

In this work, we are concerned with mixtures of non-stationary Gaussian processes of the form
\begin{equation}
    f_\mathrm{m}(\vec{x}) = \sum_i f_{i,\mathrm{warp}}(\vec{x}) \text{~~~with~} f_{i,\mathrm{warp}}\sim\mathcal{GP}(0,k_i(\phi_i(\vec{x}),\phi_i(\vec{x}')),
    \label{eq:fullGP}
\end{equation}
where $k_i$ are product separable along input dimensions and stationary (that is, $k_i(\vec{x},\vec{x}') = k_i(\vec{\tau})$, with $\vec{\tau} = \vec{x}-\vec{x}'$). The functions $\phi_i: \mathcal{D}_{\mathrm{in}} \rightarrow \mathcal{D}_i$ are invertible space warping functions with $\mathcal{D}_{\mathrm{in}}\subseteq\mathbb{R}^D$, $\mathcal{D}_i\subseteq\mathbb{R}^D$ and do not have singularities in the input domain $\mathcal{D}_{\mathrm{in}}$. The full kernel corresponding to (\ref{eq:fullGP}) is given by $k_\mathrm{m}(\vec{x},\vec{x}') = \sum_i k_{i,\mathrm{warp}}(\vec{x},\vec{x}')= \sum_i k_i(\phi_i(\vec{x}),\phi_i(\vec{x}'))$.
The kernel property would be preserved also for non-invertible space warping functions, but for reasons that will become clear later, we focus on invertible functions. One of the main difficulties in using (\ref{eq:fullGP}) in GP regression is that product separability of the kernel is lost due to the warping function. 

A particularly important use case of this model is the separation of non-stationary processes. In the GP framework, source separation can be achieved by extracting the function corresponding to the $j$th source of the mixture via the posterior $\vec{f}_{j, \mathrm{warp}}\vert \vec{y}$. Again, this posterior has a closed form solution~\cite{liutkus_2011}, which is given by equation~(\ref{eq:standard_GPR}) replacing $k_f$ with $k_{j,\mathrm{warp}}$, $k_\epsilon$ with $\sum_{i\neq j} k_{i,\mathrm{warp}}$ and choosing $X=X_*$. 

\subsection{WarpSKI}
\label{sec:warpSKI}

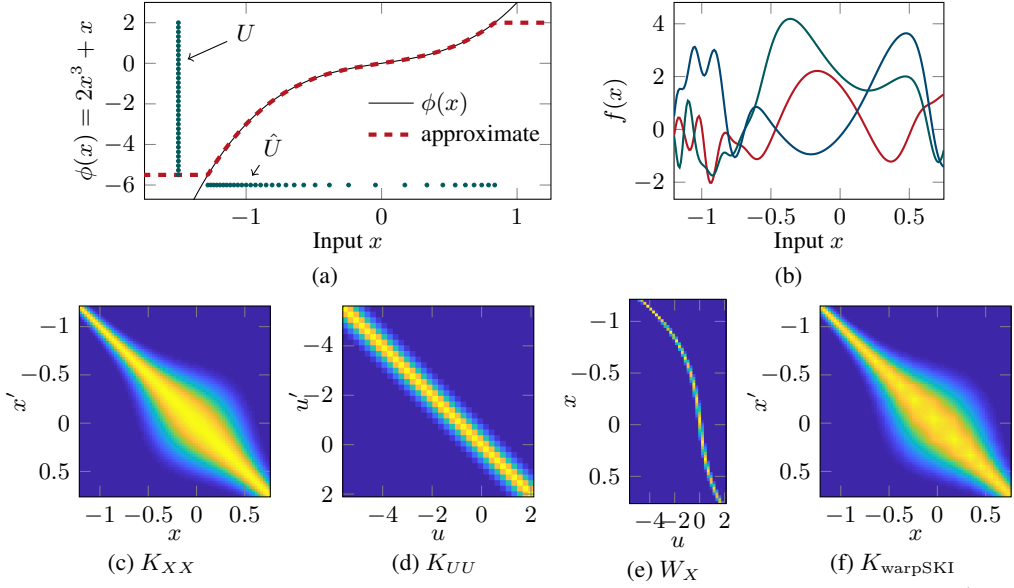
\begin{figure}[tb]
\centering
\begin{subfigure}[b]{0.5\textwidth}
    \begin{tikzpicture}[font=\footnotesize]
\begin{axis}[height=4.2cm, width=1.0\textwidth,
             xmin=-1.75, xmax=1.25,
             ymin=-6.7, ymax=3,
             xlabel = Input $x$, ylabel = {$ \phi(x) = 2x^3 + x$},
             ylabel near ticks, xlabel near ticks,
             ylabel shift = -1ex, xlabel shift = -1ex,
             legend pos=south east,
             legend cell align={left},
             legend style={draw=none,fill=none,yshift=0.5cm,xshift=0.2cm}]
\node (N1) at (axis cs:-1.5,0) {};
\node (N2) at (axis cs:-1,1.5) {$U$};
\draw[->](N2)--(N1);
\node (N3) at (axis cs:-1,-6) {};
\node (N4) at (axis cs:-0.8,-4) {$\hat U$};
\draw[->](N4)--(N3);
\addplot[mark=none, domain=-2:2, samples=200] {2*x^3 + x};
\addplot[mark=none, ultra thick, dashed, color = uzl_red1] table [x index = {0}, y index = {1}, col sep=comma] {appr_warp_func.csv};
\addplot[only marks, mark options={scale=0.35}, x filter/.code={\pgfmathparse{\pgfmathresult-1.5}}, color = uzl_green2] table[x index = {0}, y index = {1}, col sep=comma]{equidistant_inducing_points.csv};
\addplot[only marks, mark options={scale=0.35}, y filter/.code={\pgfmathparse{\pgfmathresult-6}}, color = uzl_green2] table[x index = {0}, y index = {1}, col sep=comma]{warped_inducing_points.csv};
\legend{{$\phi(x)$},approximate}
\end{axis}
\end{tikzpicture}
    \caption{}
\end{subfigure}
\begin{subfigure}[b]{0.37\textwidth}
    \begin{tikzpicture}[font=\footnotesize]
\begin{axis}[height=4.2cm, width=1.0\textwidth,
             xmin=-1.2, xmax=0.75,
             xlabel = Input $x$, ylabel = $f(x)$,
             ylabel near ticks, xlabel near ticks,
             ylabel shift = -1ex, xlabel shift = -1ex]
\addplot[mark=none, thick, color = uzl_red1] table [x index = {0}, y index = {1}, col sep=comma] {samples.csv};
\addplot[mark=none, thick, color = uzl_green2] table [x index = {0}, y index = {2}, col sep=comma] {samples.csv};
\addplot[mark=none, thick, color = uzl_blue] table [x index = {0}, y index = {3}, col sep=comma] {samples.csv};
\end{axis}
\end{tikzpicture}
    \caption{}
\end{subfigure}\\ \vspace{0.2em}

\begin{subfigure}[c]{0.295\textwidth}
\begin{tikzpicture}[font=\footnotesize]
\begin{axis}[height=4.3cm,
             width=1.0\textwidth,
             enlargelimits=false,
            axis on top,
            axis equal image,
            y dir=reverse,
            xlabel = $x$, ylabel = $x'$,
            ylabel near ticks, xlabel near ticks,
            ylabel shift = -2ex, xlabel shift = -1ex]
    \addplot graphics
    	[xmin=-1.2139,xmax=0.7639,ymin=-1.2139,ymax=0.7639]{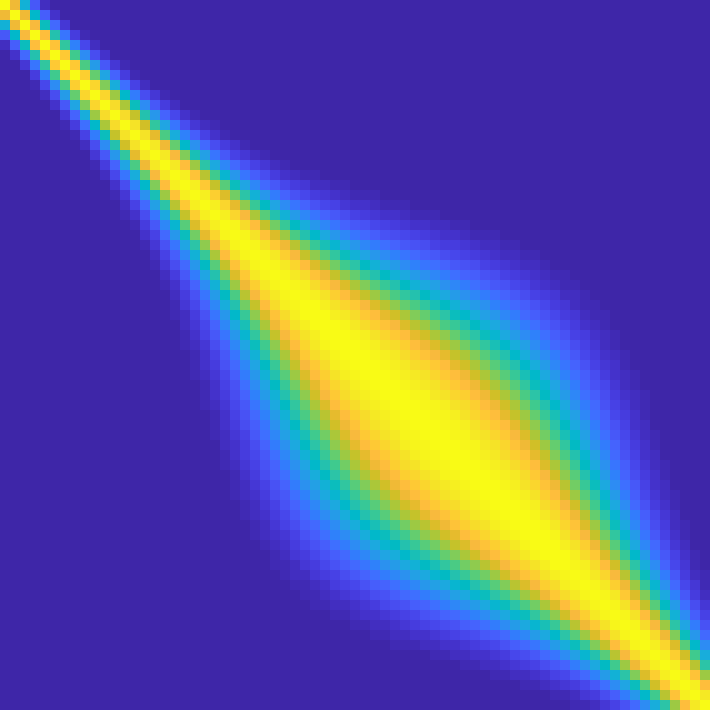};
    \end{axis}
\end{tikzpicture}
    \caption{$K_{X X}$}
\end{subfigure} \hspace{-1.5em}
\begin{subfigure}[c]{0.295\textwidth}
\begin{tikzpicture}[font=\footnotesize]
\begin{axis}[height=4.3cm,
             width=1.0\textwidth,
             enlargelimits=false,
            axis on top,
            axis equal image,
            y dir=reverse,
            xlabel = $u$, ylabel = $u'$,
            ylabel near ticks, xlabel near ticks,
            ylabel shift = -2ex, xlabel shift = -1ex]
    \addplot graphics
    	[xmin=-5.6136,xmax=2.1136,ymin=-5.6136,ymax=2.1136]{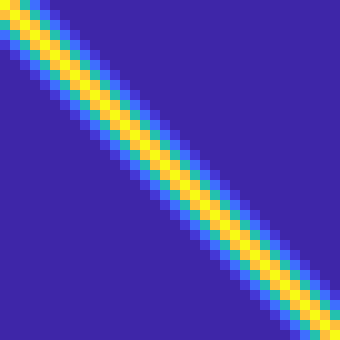};
    \end{axis}
\end{tikzpicture}
    \caption{$K_{U U}$}
\end{subfigure}\hspace{-1.5em}
\begin{subfigure}[c]{0.205\textwidth}
    \begin{tikzpicture}[font=\footnotesize]
\begin{axis}[height=4.3cm,
             width=1.0\textwidth,
             enlargelimits=false,
            axis on top,
            y dir=reverse,
            xlabel = $u$, ylabel = $x$,
            ylabel near ticks, xlabel near ticks,
            ylabel shift = -2ex, xlabel shift = -1ex
            ]
    \addplot graphics
    	[xmin=-5.6136,xmax=2.1136,ymin=-1.2139,ymax=0.7639]{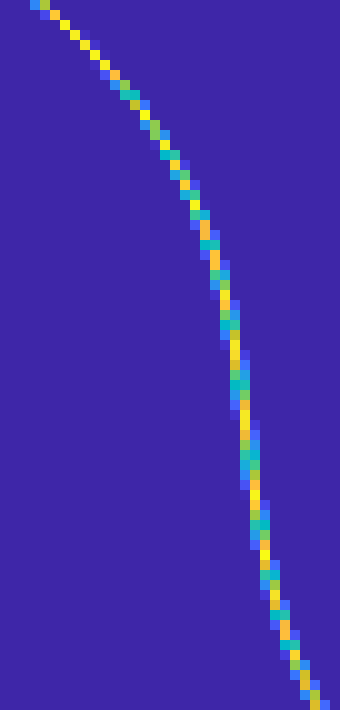};
    \end{axis}
\end{tikzpicture}
    \caption{$W_X$}
\end{subfigure}\hspace{-1.5em}
\begin{subfigure}[c]{0.295\textwidth}
\begin{tikzpicture}[font=\footnotesize]
\begin{axis}[height=4.3cm,
             width=1.0\textwidth,
             enlargelimits=false,
            axis on top,
            axis equal image,
            y dir=reverse,
            xlabel = $x$, ylabel = $x'$,
            ylabel near ticks, xlabel near ticks,
            ylabel shift = -2ex, xlabel shift = -1ex]
    \addplot graphics
    	[xmin=-1.2139,xmax=0.7639,ymin=-1.2139,ymax=0.7639]{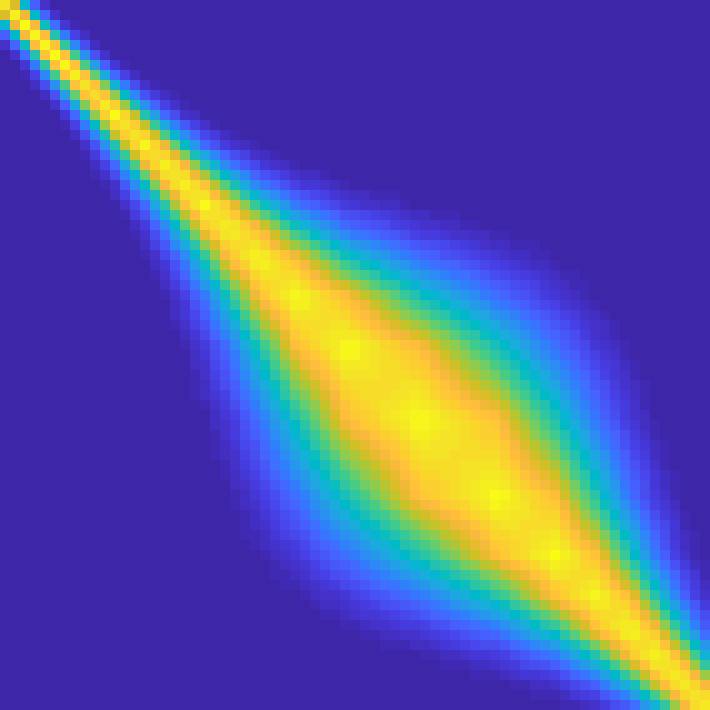};
    \end{axis}
\end{tikzpicture}
    \caption{$K_{\mathrm{warpSKI}}$}
\end{subfigure}

\caption{Illustrative example of warpSKI for squared exponential kernel $k_{\mathrm{SE}}(\phi(x),\phi(x'))$ with $\phi(x)=2x^3+x$. In (a), the warping function $\phi$ and the equidistant inducing points $U$ as well as the non-equidistant inducing points $\hat U$ are depicted. The approximate warping function induced by $U$ and $\hat U$ closely reflects the true function within the support domain of $U$ and $\hat U$. In (b), some samples from the warped kernel are depicted. Figures (c) to (f) show the different matrices involved in warpSKI.}
\label{fig:illustrative_example}
\end{figure}

We wish to reduce the time and storage costs for marginal likelihood evaluations and for the regression/source separation problem when the GP is a mixture of processes with a non-stationary phase as in (\ref{eq:fullGP}). We do not want to introduce any restrictions on the structure of input points, in particular partial grid structures or missing values must be supported. Note, that structure in kernels with a non-stationary phase cannot easily be exploited: previously proposed methods fail in this case because they either assume stationarity of the kernel \cite{Hartikainen2010KalmanFA, Solin2018InfiniteHorizonGP, Cunningham_2008} or rely on its separability \cite{Saatci_scalableinference, Wilson_2015}. To make things worse, the summation structure in (\ref{eq:fullGP}) alone leads to the loss of the Kronecker product property for the whole kernel $k_m$ even when all $\phi_i$ are linear functions. 

In the following we apply structured kernel interpolation to mixtures of non-stationary functions in an extension we call warpSKI. We propose to employ multiple sets of inducing points to recover structure in the kernel of the GP in (\ref{eq:fullGP}) and enable scalable GP regression.

As a first step we consider a single warped kernel $k_{\mathrm{warp}}(\vec{x},\vec{x}') = k(\phi(\vec{x}),\phi(\vec{x}'))$, corresponding to one of the summands in (\ref{eq:fullGP}). It is easily verified, that SKI can be applied to the stationary and separable kernel $k$ when using the warped input points $\vec{z} = \phi(\vec{x})$. This leads to an approximate form for the warped kernel matrix
\begin{equation}
    K_{\mathrm{warp}, X X} \approx W_Z K_{U U} W_Z^\top,
    \label{eq:warpSKI1}
\end{equation}
where we have used $\tilde{K}_{Z U} = W_Z K_{U U}$ and $W_Z$ interpolates between the inducing points and the warped points $Z=\{ \phi(\vec{x}_1), \dots, \phi(\vec{x}_n) \}\coloneqq \Phi(X)$. The matrix $K_{U U}$ now only depends on the stationary/separable kernel $k$ and thus structure (Kronecker and Toeplitz) can be imposed by placement of $U$. Interestingly, the non-stationarity of the warped kernel $k_{\mathrm{warp}}$ is now fully embedded in the sparse matrix $W_Z$. 

We propose a reinterpretation for the case that $\phi$ is invertible: Instead of using interpolation between the warped points $Z=\Phi(X)$ and the inducing points $U$ (placed on an equispaced rectilinear grid), one can equivalently interpolate between the original input points $X$ and a set of inducing points formed as $\hat U = \Phi^{-1}(U)$. This leads to the approximation
\begin{equation}
    K_{\mathrm{warp}, X X} \approx W_X K_{\mathrm{warp}, \hat U \hat U} W_X^\top = W_X K_{U U} W_X^\top = (\ref{eq:warpSKI1}) \coloneqq K_{\mathrm{warpSKI}},
\end{equation}
where we have used $\tilde{K}_{X \hat U} = W_X K_{\mathrm{warp}, \hat U \hat U}$ and $W_X$ interpolates between the inducing points $\hat U$ and the input points $X$. Here, in contrast to standard SKI, the inducing point set~$\hat U$ does have a warped (non-equidistant) lattice structure and thereby accounts for the warping function. Using this consideration, we now have an approximate form $ W_X K_{U U} W_X^\top$ for the warped kernel~$k_{\mathrm{warp}}$, that is fully specified in the structure of an inducing point set $\hat U$ and a stationary and separable kernel~$k$.
This can be seen as an extension to SKI, that is able to generate rich kernel structure by placing non-equidistant/warped inducing points $\hat U$. See Figure~\ref{fig:illustrative_example} as an illustrative example for all matrices involved in warpSKI.

We continue by specifying the approximate form for the full non-stationary kernel in (\ref{eq:fullGP}):
\begin{equation}
    K_{\mathrm{m}, X X} \approx \sum_i K_{i, \mathrm{warpSKI}} = \sum_i W_{i, X} K_{i, U_i U_i} W_{i, X}^{^\top},
    \label{eq:full_approx_form}
\end{equation}
where we use multiple sets of non-equispaced/warped inducing point sets $\hat U_i$ and $W_{i, X}$ interpolating between those points and the input points $X$. Note, that the space warping functions are now fully embedded in the matrices $W_{i,X}$ through placement of $\hat U_i$ and structure (Toeplitz/Kronecker) is imposed on $K_{i, U_i U_i}$ by placing $U_i$. Fast MVMs are possible and the inference problem can be solved by linear conjugate gradients requiring $j\ll n$ steps for convergence up to machine precision leading to $\mathcal{O}(n+g(m))$ runtime with $g(m)\le m\log{m}$.



\subsection{Spatio-Temporal Gaussian Processes}
Next, we consider the case, that the space warping function $\phi$ is an elementwise function, which means that it can be written as a vector of one dimensional functions
\begin{equation}
\phi(\vec{x}) = \left[ \phi^{(1)}(\vec{x}^{(1)}), \dots, \phi^{(D)}(\vec{x}^{(D)}) \right]^\top.
\end{equation}
When the space warping functions $\phi_i$ in (\ref{eq:fullGP}) are elementwise functions, the summands $k_{i,\mathrm{warp}}$ become product separable and the whole kernel can be written as
\begin{equation}
    k_\mathrm{m}(\vec{x},\vec{z}) = \sum_i \prod_d k_i^{(d)}\left(\phi_i^{(d)}(\vec{x}^{(d)}),\phi_i^{(d)}(\vec{z}^{(d)})\right).
    \label{eq:kernel_sum}
\end{equation}
An important special case of~(\ref{eq:kernel_sum}) is given by the spatio-temporal covariance function
\begin{equation}
    k_\mathrm{m}(\vec{s},\vec{s}',t,t') = \sum_i k_{i,\vec{s}}(\vec{s}, \vec{s}') k_{i,\mathrm{t}}(\phi_i(t),\phi_i(t')),
    \label{eq:spatio_temp_kernel}
\end{equation}
where non-stationarity is only assumed for the temporal domain -- in this case $\phi_i(t)$ is called a time warping function \cite{Mueller2007}. When $k_{i,\mathrm{t}}$ is a periodic kernel, $\phi_i(t)$ can be considered to be a `phase warping function' and maps from the time domain to multiples of $2\pi$, thus specifying the period length.
Such models often arise in biomedical applications due to superposition of physiological processes such as cardiac or respiratory activity, which are inherently non-stationary in time. In many practical large-scale biomedical problems, this necessitates efficient solutions to the corresponding inference and model learning task.

Note, that the kernel in (\ref{eq:kernel_sum}) enables Kronecker structure MVMs also for standard SKI (with equidistant lattice inducing points), but Toeplitz structure is lost. In contrast, warpSKI (with non-equidistant lattice inducing points) does recover Kronecker and Toeplitz structure. We argue, that Toeplitz structure is particularly important in temporal and spatio-temporal regression problems due to the possibly long time axis and that Kronecker structure exploitation can quickly become prohibitive with respect to the temporal domain. Therefore, warpSKI offers important advantages and poses no restrictions on the amount of inducing points placed over the temporal axis. A comparison between SKI and warpSKI can be found in the Appendix. 

\subsection{Fast Source Separation}
Having approximated the solution to the linear problem $\vec{\alpha} \approx \vec{\tilde \alpha} = \left[ \sum_i K_{i, \mathrm{warpSKI}} + \sigma^2 I \right]^{-1}\vec{y}$ via conjugate gradients, we continue to consider the source separation problem \cite{liutkus_2011}, i.e.\ our goal is to approximate the mean of the posterior $\vec{f}_{j, \mathrm{warp}}\vert \vec{y}$ corresponding to the $j$th source in the mixture. The standard GP solution to the posterior mean is given by $K_{j,\mathrm{warp}, X_* X}\vec{\tilde \alpha}$ and would require $\mathcal{O}(n^2)$ time for $n$ test points $X_*=X$. For the source separation problem we can again exploit kernel structure using the approximation
\begin{equation}
    \mathbb{E}\left[ \vec{f}_{j, \mathrm{warp}}\vert \vec{y} \right] \approx K_{j,\mathrm{warpSKI}} \vec{\tilde \alpha} =  W_{j, X} K_{j, U_j U_j} W_{j, X}^{^\top} \vec{\tilde{\alpha}}.
\end{equation}
The complexity for source separation, once $\vec{\tilde \alpha}$ was obtained, is $\mathcal{O}(n + m \log m)$ for Toeplitz structure in $K_{j,U_j U_j}$ and quasi-linear $\mathcal{O}(n+g(m))$ (with $g(m)\approx m$) for Kronecker/Toeplitz structure.


\subsection{Fast Model Learning}
The hyperparameters of all kernels in the sum can be learned by jointly optimizing the marginal likelihood. Different structure exploiting approximations have been proposed. Unfortunately, the model specified in (\ref{eq:fullGP}) does not allow to use the highly efficient scaled eigenvalue method \cite{Wilson_2014} due to its summation structure. Therefore, we will consider the recently introduced stochastic trace estimation approach \cite{Dong2017}, which approximates the log determinant $\mathrm{log} \, |K_{X X} + \sigma^{2}I|$ and its derivative w.r.t.\ the hyperparameters also via an iterative MVM scheme. In its core, the method uses $\mathrm{log}|\hat K| = \mathrm{trace}(\mathrm{log}(\hat K)) = \mathbb{E}[\vec{z}^\top \mathrm{log}(\hat K) \vec{z}]$, where $\vec{z}$ is commonly chosen to be a vector with Rademacher random variables as entries. Usually, few probe vectors suffice to get a good approximation -- leaving us with the task of calculating $\mathrm{log}(\hat K) \vec{z}$. For this task, a Lanczos decomposition approach has been proposed in \cite{Dong2017}. The stochastic trace estimation approach accesses the kernel matrix only through MVMs and is thus compatible with the proposed approximate form in~(\ref{eq:full_approx_form}).


\section{Experiments}
We evaluate the proposed non-stationary GP framework on a numerical dataset and then we aim to motivate the usefulness of the method in relevant large-scale biomedical applications. One of the examples concerns the extraction of fetal ECG signals from abdominal recordings -- GPs were suggested in the literature before in this context but large datasets could not be processed so far \cite{Niknazar2012}. Another example treats the separation of perfusion and ventilation related effects in electrical impedance tomography (EIT) images of the lung, which is a long-standing problem with important implications in mechanical ventilation \cite{Pikkemaat_2010}. Many previous approaches to perfusion/ventilation separation are not satisfactory. To the best of our knowledge, GPs were not considered in this context before -- the corresponding problem becomes solvable through the methods proposed in this paper.

WarpSKI was implemented in MATLAB as an extension to the GPML~4.2 library, all experiments were carried out on a workstation with an INTEL Core i7-6700K CPU. In all experiments, \mbox{L-BFGS}~\cite{Liu1989} was used for hyperparameter learning, respectively with a maximum of 100 optimization steps.

\subsection{Numerical Data}
As a first test, we apply warpSKI to a mildly non-linear separable warping function on a numerical 2D example. Samples are generated from a warped squared exponential kernel of the form $k_{\mathrm{SE}} (\phi(\vec{x}),\phi(\vec{x})')$, where $\phi$ is given by $\phi([x_1, x_2]^\top)=[ 2x_1^3+x_1,x_2]^\top$ and the hyperparameters were set to $\ell_{\mathrm{SE}}=0.4$ and $\sigma_{\mathrm{SE}}=1.5$. The input point positions are sampled from a uniform distribution within a rectangular area (spanning $[-1.2,0.75]\times[-2.5,2.5]\subset\mathbb{R}^2$). 
To account for the non-stationarity of the kernel, one non-equidistant inducing point set $\hat U$ is used, as described in Section~(\ref{sec:warpSKI}).

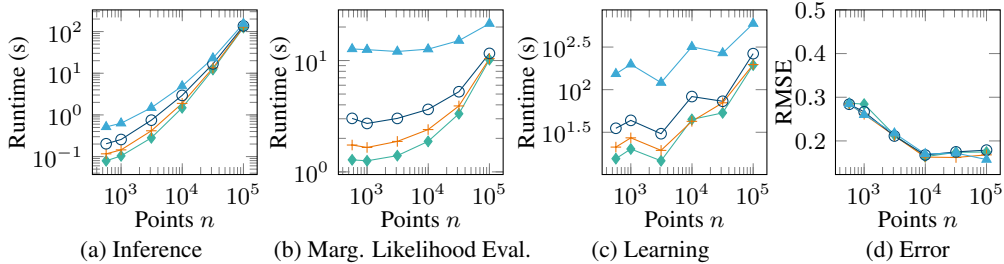
\begin{figure}[h!]
    \centering
    \begin{subfigure}[b]{0.27\textwidth}
        \begin{tikzpicture}[font=\footnotesize]
    \begin{loglogaxis}[height=1.0\textwidth, width=1.0\textwidth,
             xlabel = Points $n$, ylabel = Runtime (s),
             ylabel near ticks, xlabel near ticks,
             ylabel shift = -1.25ex, xlabel shift = -1ex]
         \addplot[mark=diamond*, color = uzl_green_light] table [x index = {0}, y index = {2}, col sep=comma] {inference_times.csv};
        \addplot[mark=+, color = uzl_orange] table [x index = {0}, y index = {3}, col sep=comma] {inference_times.csv};
        \addplot[mark=o, color = uzl_blue] table [x index = {0}, y index = {4}, col sep=comma] {inference_times.csv};
        \addplot[mark=triangle*, color = uzl_blue_light] table [x index = {0}, y index = {5}, col sep=comma] {inference_times.csv};
    \end{loglogaxis}
\end{tikzpicture}
    \caption{Inference}
    \end{subfigure} \hspace{-1.35em}
    \begin{subfigure}[b]{0.27\textwidth}
        \begin{tikzpicture}[font=\footnotesize]
\begin{loglogaxis}[height=1.0\textwidth, width=1.0\textwidth,
         xlabel = Points $n$, ylabel = Runtime (s),
         ylabel near ticks, xlabel near ticks,
         ylabel shift = -1.25ex, xlabel shift = -1ex]
     \addplot[mark=diamond*, color = uzl_green_light] table [x index = {0}, y index = {2}, col sep=comma] {marg_lik_eval_times.csv};
    \addplot[mark=+, color = uzl_orange] table [x index = {0}, y index = {3}, col sep=comma] {marg_lik_eval_times.csv};
    \addplot[mark=o, color = uzl_blue] table [x index = {0}, y index = {4}, col sep=comma] {marg_lik_eval_times.csv};
    \addplot[mark=triangle*, color = uzl_blue_light] table [x index = {0}, y index = {5}, col sep=comma] {marg_lik_eval_times.csv};
\end{loglogaxis}
\end{tikzpicture}
    \caption{Marg. Likelihood Eval.}
    \end{subfigure} \hspace{-1.85em}
    \begin{subfigure}[b]{0.27\textwidth}
        \begin{tikzpicture}[font=\footnotesize]
\begin{loglogaxis}[height=1.0\textwidth, width=1.0\textwidth,
         xlabel = Points $n$, ylabel = Runtime (s),
         ylabel near ticks, xlabel near ticks,
         ylabel shift = -1.25ex, xlabel shift = -1ex, ytick={31.62,100,316.23}]
     \addplot[mark=diamond*, color = uzl_green_light] table [x index = {0}, y index = {2}, col sep=comma] {learning_times.csv};
    \addplot[mark=+, color = uzl_orange] table [x index = {0}, y index = {3}, col sep=comma] {learning_times.csv};
    \addplot[mark=o, color = uzl_blue] table [x index = {0}, y index = {4}, col sep=comma] {learning_times.csv};
    \addplot[mark=triangle*, color = uzl_blue_light] table [x index = {0}, y index = {5}, col sep=comma] {learning_times.csv};
\end{loglogaxis}
\end{tikzpicture}
    \caption{Learning}
    \end{subfigure} \hspace{-1.5em}
    \begin{subfigure}[b]{0.27\textwidth}
        \begin{tikzpicture}[font=\footnotesize]
\begin{axis}[height=1.0\textwidth, width=1.0\textwidth,
         xlabel = Points $n$, ylabel = RMSE,
         ylabel near ticks, xlabel near ticks,
         ylabel shift = -1.25ex, xlabel shift = -1ex,
         xmode=log, ymax=0.5]
     \addplot[mark=diamond*, color = uzl_green_light] table [x index = {0}, y index = {2}, col sep=comma] {regression_errors.csv};
    \addplot[mark=+, color = uzl_orange] table [x index = {0}, y index = {3}, col sep=comma] {regression_errors.csv};
    \addplot[mark=o, color = uzl_blue] table [x index = {0}, y index = {4}, col sep=comma] {regression_errors.csv};
    \addplot[mark=triangle*, color = uzl_blue_light] table [x index = {0}, y index = {5}, col sep=comma] {regression_errors.csv};
\end{axis}
\end{tikzpicture}
    \caption{Error}
    \end{subfigure}
    \caption{Results of warpSKI for numerical data with variable numbers of inducing/input points: Inference time is in (a), time for marginal likelihood evaluations is in (b), hyperparameter learning runtime in (c) and RMSE against true GP sample in (d). The number of inducing points is $m=9933$~(\markerdiamond), $m=20\,020$ (\textcolor{uzl_orange}{+}), $m=49\,824$ (\markercircle) and $m=99\,960$ (\markertriangle).}
    \label{fig:numerical_example}
\end{figure}

The generation of high-dimensional GP samples is itself a non-trivial task as it requires Cholesky decomposition of the full kernel matrix. Therefore, to generate samples with up to $10^5$ input points, we exploit the Kronecker structure of the inducing point kernel matrix and use a high number of inducing points to warp the samples with high accuracy. We then apply additive Gaussian white noise $\epsilon\sim\mathcal{N}(0,\sigma^2)$ with $\sigma =0.5$ to form the final regression targets $\vec{y}$.
In Figure~\ref{fig:numerical_example}, we show the inference and model optimization times (here, $\sigma$, $\sigma_{\mathrm{SE}}$ and $\ell_{\mathrm{SE}}$ were learned from the simulated data) as well as the root-mean-square-error (RMSE) for different input and inducing point sizes. The tolerance for conjugate gradients was set to $10^{-1}$, and marginal likelihood evaluations were done using 20 probe vectors in the stochastic trace estimation. A maximum of 100 steps was allowed for L-BFGS during the hyperparameter learning.

As this example does not include mixtures of non-stationary warping functions, it enables direct comparison of the proposed method to previous publications because it could be transformed back to an equivalent standard SKI task 
as discussed in Section~\ref{sec:warpSKI}. Therefore the results obtained for this space-warped GP match earlier results, compare for instance to \cite{Wilson_2015}. In particular, note that warpSKI/SKI is inexpensive with respect to the number of inducing points.

\subsection{Fetal ECG Data}
Next, we test the proposed fast source separation algorithm based on warpSKI on large biomedical datasets. The application of GPs to biomedical data has been proposed by many authors, but scalability to large datasets has been lacking. In this example we apply GP models to ECG data, specifically we treat the separation of fetal ECG and maternal ECG signals in baseline-free abdominal recordings. This was previously demonstrated on small datasets in \cite{Niknazar2012}, where a model as in (\ref{eq:fullGP}) with two nonlinear warping functions was used and the source separation was solved via the batch GP formulation. We first validate the proposed warpSKI on a short segment of data taken from the Physionet fetal ECG database \cite{Jezewski2012} (using the 4th channel of subject R01) and then demonstrate scalability of warpSKI to signals consisting of up to $10^5$ datapoints on a longer segment of the same data. 

Similar to \cite{Niknazar2012}, the kernel is chosen as a mixture of two phase-warped processes, i.e. $k_\mathrm{m}(t, t') = k_\mathrm{maternal}(\phi_1(t),\phi_1(t')) + k_\mathrm{fetal}(\phi_2(t),\phi_2(t'))$, where both $k_\mathrm{maternal}$ and $ k_\mathrm{fetal}$ are quasi-periodic kernels as defined in \cite{Rasmussen_2005}. Note, that standard SKI cannot be applied in this example as it does not recover Toeplitz structure.

Following \cite{Niknazar2012}, the optimization of hyperparameters in the ECG separation problem should be guided by prior knowledge by either fixing hyperparameters to reasonable estimates or by using strong hyperpriors. In this experiment, we used fixed values for the length scales of the two quasi-periodic kernel functions (which appears to be beneficial for ECG signals and is in line with \cite{Niknazar2012}) and optimize for the variance hyperparameters $\sigma_\mathrm{maternal}$ and $\sigma_\mathrm{fetal}$. In the considered dataset, the nonlinear warping functions can be extracted directly from the data by detecting fetal and maternal R~peaks in the provided reference signals and assuming a constant phase between respectively two R~peaks. In \cite{Niknazar2012}, it was also proposed to learn the full warping function by optimization of the model likelihood, which however is difficult due to the high number of local minima. The optimization of non-stationary warping functions was discussed in more detail in e.g.~\cite{pmlr-v51-heinonen16} but shall not be the focus of this work. Generally, we recommend to use reference signals for the phase whenever possible or at least use strong priors on the phase function.

\begin{figure}
    \centering
    \begin{subfigure}[b]{0.5\textwidth}
    \begin{tikzpicture}[font=\footnotesize]
        \begin{groupplot}[group style={group size=1 by 3,vertical sep=0.12cm},
                     height=2.6cm, width=1.0\textwidth,
                     ylabel near ticks, xlabel near ticks,
                     ylabel shift = -1ex, xlabel shift = -1ex,
                     ymin=-55,ymax=55,
                     enlargelimits=false,
                     axis x line=bottom,
		             axis y line=left,
		             ytick={-45,0,45}]
        \nextgroupplot[ylabel= Raw, xticklabels=\empty]
        \addplot[smooth, mark=none, color = black] table [x index = {0}, y index = {1}, col sep=comma] {ecg_excerpt.csv};
        \nextgroupplot[ylabel= $f_{\mathrm{maternal}}$, xticklabels=\empty]
        \addplot[mark=none, color = black] table [x index = {0}, y index = {2}, col sep=comma] {ecg_excerpt.csv};
        \nextgroupplot[ylabel= $f_{\mathrm{fetal}}$, xlabel=Time (s)]
        \addplot[mark=none, color = black] table [x index = {0}, y index = {3}, col sep=comma] {ecg_excerpt.csv};
        \end{groupplot}
    \end{tikzpicture}
    \caption{}
    \end{subfigure}
    \hspace{0.5cm}
    \begin{subfigure}[b]{0.4\textwidth}
    \begin{tikzpicture}[font=\footnotesize]
        \begin{axis}[height=4.2cm, width=1.0\textwidth,
                     xlabel = $\sigma_{\mathrm{maternal}}$, ylabel = $-\mathrm{log}\,\mathcal{L}(\hyp\vert\vec{y})$,
                     ylabel near ticks, xlabel near ticks,
                     ylabel shift = -1ex, xlabel shift = -1ex,
                     xmode=log, enlarge x limits=0,
                     legend pos=north east,
                     legend cell align={left},
                     legend style={draw=none,fill=none,yshift=0cm,xshift=0cm}]
        \addplot[mark=none, thick, color = black] table [x index = {0}, y index = {1}, col sep=comma] {ecg_log_lik.csv};
        \addplot[mark=none, thick, color = uzl_red1] table [x index = {0}, y index = {2}, col sep=comma] {ecg_log_lik.csv};
        \legend{Batch GP, warpSKI}
        \end{axis}
    \end{tikzpicture}
    \caption{}
    \end{subfigure}
    \caption{Results of fetal ECG extraction. In (a), an excerpt from the large-scale (100\,s) source separation is depicted. In (b), a comparison between exact and approximate warpSKI marginal likelihoods is shown on the smaller dataset (10\,s) for different values of $\sigma_{\mathrm{maternal}}$.}
    \label{fig:fetal_Ecg}
\end{figure}
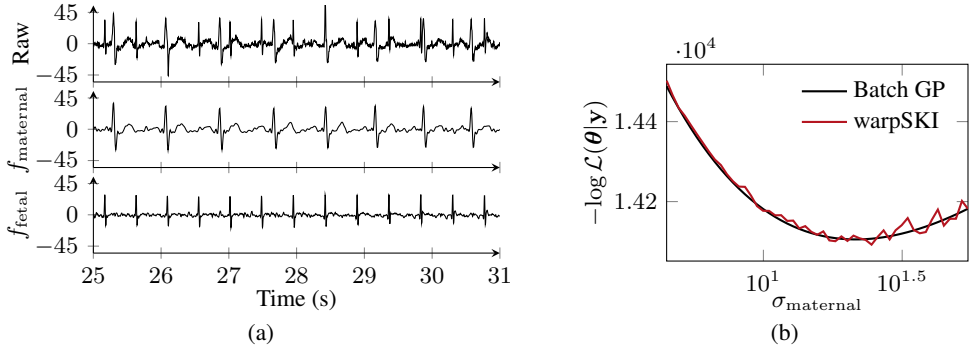

As a first step, we apply the proposed warp-SKI method to a subset of the data consisting of $n= 5000$ points (corresponding to 10 seconds of data sampled at 500\,Hz) and validate it against the batch GP solution. We then do a large-scale stress test using $10^5$ input points (corresponding to 100 seconds of data sampled at 1000\,Hz). As a measure for the separation success we use the SNR
improvement metric that was proposed in the context of ECG denoising in \cite{Bartolo1996}. For the stresstest, again 20~probe vectors were used for stochastic trace estimation and a maximum of 100 L-BFGS steps. LCG tolerance was set to $10^{-1}$ and $5\cdot 10^{-3}$ for parameter learning and source separation, respectively. In Figure~\ref{fig:fetal_Ecg}, an excerpt of the large-scale source separation as well as a comparison of exact/approximate marginal likelihood evaluations is depicted.

\begin{table}[h]
 \caption{Performance comparison of fetal ECG extraction for different input sizes and methods.}
  \centering
  \begin{tabular}{l|ccccccc}
    \toprule
     & \makecell{Input \\points} & \makecell{Inducing \\points} & \makecell{Time for \\Inference} & \makecell{Time for \\Learning} & $\sigma_\mathrm{fetal}$ & $\sigma_\mathrm{maternal}$ & \makecell{SNR \\Improvement}\\
    \midrule
    full GP & 5000 & -- & 2.60\,s & 28.5\,s & 5.08 & 21.48 & 18.6\,dB \\
    warpSKI & 5000 & \makecell{3400 (m)\\+4800 (f)} & 0.27\,s & 47.4\,s & 5.95 & 21.42 & 18.1\,dB \\
    warpSKI & 100\,000 & \makecell{14\,300 (m)\\+21\,600 (f)} & 4.14\,s & 462.3\,s & 5.11 & 32.11 & 18.2\,dB \\
    \bottomrule
  \end{tabular}
  \label{tab:fetal_Ecg}
\end{table}

\subsection{Electrical Impedance Tomography Data}
As a last example we consider a non-stationary spatio-temporal signal separation problem given by electrical impedance tomography (EIT) images of the chest. In this example EIT is used to measure regional changes in the impedance of the lung, which are caused either by changes in the ventilation or perfusion of the lung tissue. The separation of these two effects is a long-standing problem, previous approaches applied pixel-wise Fourier filtering, which however omits the spatial structure of pixels and cannot fully separate the two effects due to significant overlap in the spectrum \cite{Pikkemaat_2010}.
We show that the separation of the two pulsatile components in EIT images can be posed as a spatio-temporal GP regression problem using a mixture of non-stationary kernels. Inference and model learning can then be solved efficiently via the methods proposed in this paper.

As a model for the two superposed effects we use the spatio-temporal kernel $k_\mathrm{m}(\vec{s},\vec{s}', t,t') = k_\mathrm{vent, SE}(\vec{s},\vec{s}')k_\mathrm{vent,QP}(\phi_1(t),\phi_1(t')) + k_\mathrm{perf, SE}(\vec{s},\vec{s}')k_\mathrm{perf,QP}(\phi_2(t),\phi_2(t'))$, where a squared exponential kernel is assumed for the spatial domain and quasi-periodicity for the temporal domain in both signal components (ventilation and perfusion).

The considered dataset of a spontaneously breathing neonate is taken from \cite{Heinrich2006}. Note, that the EIT problem has `partial-grid' structure, consisting only of pixels within a circular area.  We use the first 215 frames to train our model and as a measure of training success we predict the next frame and evaluate the prediction error (using normalized root-mean-square error). As in the previous example, the phase warping function is determined directly from the data – the respiratory phase was extracted from a pixel belonging to the left lung, the cardiac phase was extracted from a pixel between the two lungs.
The total number of input points is $n=699\,825$, for warpSKI, two sets of non-equidistant inducing points with respectively 3~million and 2~million points are used. The problem is thus among the largest problems ever considered in the GP literature. Note, that standard SKI could also be applied for each of the kernels in the sum but, in contrast to warpSKI, would not recover Toeplitz structure and thus does not enable scalability to longer recordings.

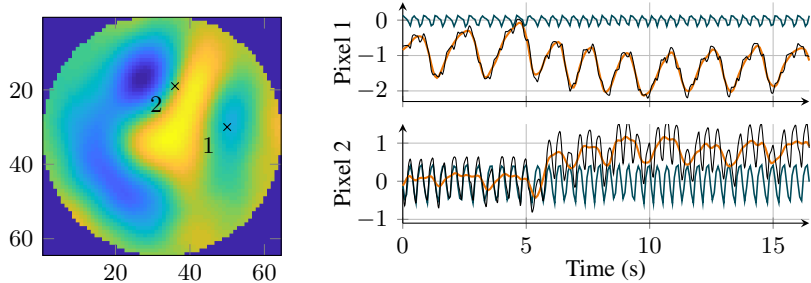
\begin{figure}[tb]
    \centering
    \begin{tikzpicture}[font=\footnotesize]
    \begin{axis}[height=4.75cm,
                 width=4.75cm,
                 enlargelimits=false,
                axis on top,
                axis equal image,
                y dir=reverse]
        \addplot graphics [xmin=0.5,xmax=64.5,ymin=0.5,ymax=64.5]{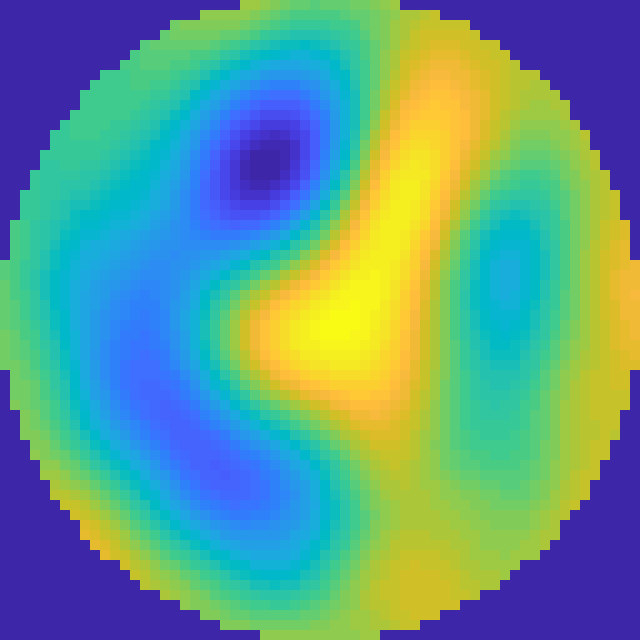};
        \addplot[color=black, mark=x, only marks] coordinates {
            (36,19)
            (50,30)};
        \node (N1) at (axis cs:31,24) {$2$};
        \node (N1) at (axis cs:45,35) {$1$};
    \end{axis}
    \end{tikzpicture} \hspace{1em}
    \begin{tikzpicture}[font=\footnotesize]
        \begin{groupplot}[group style={group size=1 by 2,vertical sep=0.3cm},
                     height=2.9cm, width=0.5\textwidth,
                     ylabel near ticks, xlabel near ticks,
                     ylabel shift = -1ex, xlabel shift = -1ex,
                     enlarge x limits=0,
                     grid=major,
                     axis x line=bottom,
		             axis y line=left]
        \nextgroupplot[ylabel= Pixel 1, xticklabels=\empty, ymin=-2.3, ymax=0.5]
        
        \addplot[mark=none, thick, opacity = 1, color = uzl_orange] table [x index = {0}, y index = {2}, col sep=comma] {eit_traces.csv};
        \addplot[mark=none, semithick, opacity = 1, color = uzl_green] table [x index = {0}, y index = {3}, col sep=comma] {eit_traces.csv};
        \addplot[smooth, mark=none, color = black] table [x index = {0}, y index = {1}, col sep=comma] {eit_traces.csv};
        
        \nextgroupplot[ylabel= Pixel 2, xlabel=Time (s), ymin=-1.1, ymax=1.5]
        
        \addplot[mark=none, semithick, opacity = 1, color = uzl_green] table [x index = {0}, y index = {9}, col sep=comma] {eit_traces.csv};
        \addplot[mark=none, thick, opacity = 1, color = uzl_orange] table [x index = {0}, y index = {8}, col sep=comma] {eit_traces.csv};
        \addplot[smooth, mark=none, color = black] table [x index = {0}, y index = {7}, col sep=comma] {eit_traces.csv};
        \end{groupplot}
    \end{tikzpicture}
    \caption{Result of EIT perfusion-ventilation separation. Time traces correspond to the marked pixels and include measured signals (\textbf{black}), the posterior mean of perfusion related signals (\textbf{\textcolor{uzl_green}{green}}) and the posterior mean of ventilation related signals (\textbf{\textcolor{uzl_orange}{orange}}).}
    \label{fig:EIT}
\end{figure}

As in the previous example, it is beneficial to use hyperpriors and fix some of the hyperparameters (based on prior knowledge about the data) to guide the optimization. Here, we optimize for the variances $\sigma_{\mathrm{vent}}$, $\sigma_{\mathrm{perf}}$, the length-scales of the spatial kernel $\ell_{\mathrm{vent,SE}}$, $\ell_{\mathrm{perf,SE}}$ and the length scales of the time domain $\ell_{\mathrm{vent,SE\text{-}QP}}$, $\ell_{\mathrm{perf,SE\text{-}QP}}$ with regularizing lognormal hyperpriors on all of the length-scales. 
The remaining hyperparameters $\sigma$, $\ell_{\mathrm{vent,PE\text{-}QP}}$ and $\ell_{\mathrm{perf,PE\text{-}QP}}$ were fixed based on features we already found in the data a priori. 
For this experiment only 15 probe vectors were used for stochastic trace estimation to speed up the optimization. The LCG tolerance was set to 0.25 and $10^{-2}$ for optimization and source separation, respectively. Figure~\ref{fig:EIT} shows the result of the source separation on the considered dataset.

\begin{table}[h]
 \caption{Runtime and performance of warpSKI for the EIT ventilation/perfusion separation.}
  \centering
  \begin{tabular}{l|ccccc}
    \toprule
     & \makecell{Input \\points} & \makecell{Inducing \\points} & \makecell{Time for \\Inference} & \makecell{Time for \\Learning} & \makecell{nRMSE\\on test frame}\\
    \midrule
    warpSKI & 699\,825 & \makecell{\textasciitilde 3\,million (perf)\\+ \textasciitilde 2\,million (vent)} & 159.1\,s & \textasciitilde 9\,h & 0.176 \\
    \bottomrule
  \end{tabular}
  \label{tab:table}
\end{table}

\section{Discussion}
We have extended Gaussian process structured kernel interpolation to kernels with a non-stationary phase in an approach we call warpSKI. Our approach exploits matrix structure using multiple sets of non-equidistant/warped inducing point sets. We have shown, that this allows to solve large-scale regression and source separation problems, which often arise in biomedical applications due to superposition of non-stationary physiological processes (such as respiratory/cardiac activity). In many biomedical modalities, where GPs could not by applied so far, the proposed method has a high potential of uncovering new and relevant structure in the data. 

Beyond that, we argue that the placement of non-equidistant inducing points could be generally used as a tool to account for non-stationarity and build rich kernel structure -- this idea might also be extended to other GP frameworks that are compatible with SKI/warpSKI such as \cite{Hensman_2013}. 

We see our work as part of a larger push in the recent GP literature that aims to access the kernel matrix only through matrix multiplications \cite{Gardner2018, Wang2019} thus enabling highly scalable inference and learning. Exploiting intricate matrix structure for fast matrix multiplications will be key to solving large-scale problems via GPs in the future.


\printbibliography

\clearpage
\appendix
\section{Appendix}
\subsection{Overview of Kronecker and Toeplitz Methods}
\begin{table}[h]
  \caption{Structure exploiting inference and learning methods. All kernels are assumed to be stationary and $\hat K = K + \sigma^2 I$.}
  \label{sample-table}
  \centering
  \resizebox{\textwidth}{!}{%
  \begin{tabular}{p{4.45cm}|l|l|l}
  
    \toprule
    Kernel and Inputs     & Matrix     & Linear Solve & Log Determinant \\
    
    \midrule
    Kernel is separable:  & $K=\bigotimes_{d=1}^D K_d$  & noise-free:  & noise-free:    \\
    $k(\vec{x}, \vec{z}')=\prod_{d=1}^D k_d(\vec{x}^{(d)},\vec{z}^{(d)})$ & and & $K^{-1}\vec{y}=\bigotimes_{d=1}^D K_d^{-1} \vec{y}$ & $\log \vert K\vert = \sum_i V_{i,i}$ \\
    
    \cmidrule(r){3-4}
    Inputs on a rectilinear grid:& $K=Q V Q^\top$ & noisy: &  noisy:\\
    $X=\mathcal{X}_1\times\cdots\times\mathcal{X}_D$ & &  $\hat K^{-1}\vec{y} = Q(V+\sigma^2 I)^{-1}Q^\top \vec{y}$ & $\log\vert\hat{K}\vert = \sum_i\left( V_{i,i} + \sigma^2\right)$\\
    
    \midrule
    Kernel: $k(x,x')$  & $K$ is Toeplitz  & LCG with fast MVMs &  (1) circulant approx.\ \cite{WilsonDN15}\\
    Inputs: $x\in\mathbb{R}$ and equispaced & & & (2) stoch.\ trace estim.\ \cite{Dong2017}\\
    
    \midrule
    Kernel is separable & $K\approx WK_{U U}W^\top$ & LCG with fast MVMs & (1) scaled eigenvalues \cite{WilsonDN15} \\
    Inputs unstructured & \cite{Wilson_2015} & & (2) stoch.\ trace estim.\ \cite{Dong2017}\\
    
    \bottomrule
  \end{tabular}}
\end{table}

\subsection{Structure Exploitation for Kernels with a Non-Stationary Phase}
\begin{table}[h]
  \caption{Comparison of SKI \cite{Wilson_2015} and warpSKI. Inputs may be unstructured (or have partial grid structure). The kernels $k$ (and $k_i$) are assumed to be stationary and separable. The functions $\phi_i:\mathcal{D}\rightarrow\mathcal{D}_i$ and $\phi:\mathcal{D}\rightarrow\mathcal{D}_1$ with $\mathcal{D}_{\mathrm{in}}\subseteq\mathbb{R}^D$, $\mathcal{D}_i\subseteq\mathbb{R}^D$ are invertible functions. The linear solve $\hat K^{-1}\vec{y}$ is done by conjugate gradients and the log determinant is approximated using stochastic trace estimation.}
  \label{sample-table}
  \centering
   \resizebox{\textwidth}{!}{%
  \begin{tabular}{l|c|l}
    \toprule
    Kernel & SKI \cite{Wilson_2015} recovers... & warpSKI recovers... \\
    
    \midrule
    $k(\phi(x),\phi(x'))$  & --  & Toeplitz structure\\
    with $x\in\mathbb{R}$ & &  \\
    
    \midrule
    $\sum_i k_i(\phi_i(x),\phi_i(x'))$ & -- & sum over Toeplitz structures\\
    with $x\in\mathbb{R}$ &  &  \\
    
    \midrule
    $k(\phi(\vec{x}), \phi(\vec{x}'))$, & -- & Kronecker and Toeplitz structure \\
    $\vec{x}\in\mathbb{R}^D$ and $\phi$ is not an elementwise fnc.\ & &  \\
    
    \midrule
    $k(\phi(\vec{x}), \phi(\vec{x}'))$ & Kronecker structure & Kronecker and Toeplitz structure\\
    $\vec{x}\in\mathbb{R}^D$ and $\phi$ is an elementwise fnc.\ & & \\
    
    \midrule
    $\sum_i k_i(\phi_i(\vec{x}), \phi_i(\vec{x}'))$ & \multicolumn{1}{l|}{sum over} & sum over Kronecker and Toeplitz\\
    $\vec{x}\in\mathbb{R}^D$ and $\phi_i$ are elementwise fnc.\ & Kronecker structures  & structures\\
    
    \bottomrule
  \end{tabular}
  }
\end{table}

\end{document}